\documentclass[11pt,a4paper]{article}

\usepackage[table]{xcolor} %color in tables 
\usepackage{pgfplots}
\pgfplotsset{compat=1.7}

\usepackage[hyperref]{acl2017}
\usepackage{times}
\usepackage{latexsym}
\usepackage{xcolor,soul}

\usepackage{url}

\usepackage{slantsc}% http://ctan.org/pkg/slantsc
\usepackage{caption} 
\usepackage{csvsimple} %read csv file and import in a table.
\usepackage{amsmath}
\usepackage{multirow}
\usepackage{bm} %put bolded text in formula
\usepackage[utf8]{inputenc}
\usepackage{tabularx} %fit the table to column width
\usepackage[inline]{enumitem}   %inline enumerator
\usepackage{xspace} % add space after a command
\usepackage{amssymb} % adding checkmark

\aclfinalcopy % Uncomment this line for the final submission
\def\aclpaperid{***} %  Enter the acl Paper ID here

\def \dict {ConcoLeDisCo}

\definecolor{LightGray}{gray}{0.85}
\definecolor{Gray}{gray}{0.75}

\title{Automatic Mapping of French Discourse Connectives \\ to PDTB Discourse Relations}

\author{Majid Laali \hspace{2cm} Leila Kosseim\\
  Department of Computer Science and Software Engineering \\
  Concordia University, Montreal, Quebec, Canada \\
  {\tt \{m\_laali, kosseim\}@encs.concordia.ca} \\
  }
\date{}

\begin{document}

\maketitle

\begin{abstract}
   In this paper, we present an approach to exploit phrase tables generated by statistical machine translation in order to map French discourse connectives to discourse relations. Using this approach, we created \dict, a lexicon of French discourse connectives and their PDTB relations. When evaluated against  LEXCONN, \dict\ achieves a recall of 0.81 and an Average Precision of 0.68 for the \textsc{Concession} and \textsc{Condition} relations.

\end{abstract}

\section{Introduction}

Discourse connectives (DCs) (e.g. \textit{because}, \textit{although}) are terms that explicitly signal discourse relations within a text. Building a lexicon of DCs, where each connective is mapped to the discourse relations it can signal, is not an easy task. To build such lexicons, it is necessary to have linguists manually analyse the usage of individual DCs through a corpus study, which is an expensive endeavour both in terms of time and expertise. For example, LEXCONN~\cite{roze12}, a manually built lexicon of French DCs, was initiated in 2010 and released its first edition in 2012. The latest version, LEXCONN V2.1~\cite{danlos15}, contains 343 DCs mapped to an average of 1.3 discourse relations. This project is still ongoing as 37 DCs still have not been assigned to any discourse relation. Because of this, only a limited number of languages currently possess such lexicons (e.g. French~\cite{roze12}, Spanish~\cite{alonsoalemany02}, German~\cite{stede98}).  

In this paper, we propose an approach to automatically map French DCs to their associated PDTB discourse relations using parallel texts. Our approach can also automatically identify the usage of a DC where the DC signals a specific discourse relation. This can help linguists to study a DC in parallel texts and/or to find evidence for an association between discourse relations and DCs. Our approach is based on phrase tables generated by statistical machine translation and makes no assumption about the target language except the availability of a parallel corpus with another language for which a discourse parser exists; hence the approach is easy to expand to other languages. 
%In our work, we assume that the list of DCs is given or generated automatically, for example, using the approach described in~\cite{laali14}. 

We applied our approach to the Europarl corpus~\cite{koehn05} and generated \dict\footnote{\dict~is publicly available at \url{https://github.com/mjlaali/ConcoLeDisCo}.}, a lexicon mapping French DCs to their associated Penn Discourse Treebank (PDTB) discourse relations \cite{prasad08}. To our knowledge, \dict\ is the first lexicon of French discourse connectives mapped to the PDTB relation set. When compared to LEXCONN, \dict\ achieves a recall of 0.81 and an Average Precision of 0.68 for the \textsc{Concession} and \textsc{Condition} discourse relations.
 
%We chose these two discourse relations because LEXCONN uses a different set of discourse relations than the PDTB. Therefore, we picked the relations which are common in these inventories.

\section{Related Work}
\label{sec:related-work}

Lexicons of DCs have been developed for several languages: English~\cite{knott96}, Spanish~\cite{alonsoalemany02}, German~\cite{stede98}, Czech~\cite{polakova13}, and French~\cite{roze12}. However, constructing such lexicons requires linguistic expertise and is a time-consuming task. 
%For example, to build LEXCONN, \citet{roze12} gathered a list of 600 potential DCs, then, manually applied syntactic, semantic, and discourse tests to filter this initial list and identify true DCs. Finally, through an exhaustive manual analysis, they associated each connective to the discourse relations they can signal. 

Discourse connectives and their translations have been studied within parallel texts by many \cite{meyer11,meyer11-a,taboada12,cartoni13,zufferey14,zufferey14-a,zufferey15,hoek15}. These works have either focused on the effect of the translation of discourse connectives on machine translation systems \cite{meyer11,meyer11-a,cartoni13} or on a small number of discourse connectives due to the cost of manual annotations \cite{taboada12,zufferey14,zufferey14-a,zufferey15,hoek15}.

To our knowledge, very little research has addressed the automatic construction of lexicons of DCs. \citet{hidey16} proposed an automatic approach to identify English expressions that signal the \textsc{Causal} discourse relation. On the other hand, \citet{laali14} automatically extracted French DCs from parallel texts; however, they did not associate discourse relations to the extracted DCs. The proposed approach goes beyond this work by mapping DCs to their associated discourse relations. 

\section{Methodology}
\label{sec:method}

\subsection{Corpus Preparation}
For our experiments, we used the English-French part of Europarl~\cite{koehn05} which contains 2 million\footnote{2,007,723 to be exact.} parallel sentences. 
%The Europarl corpus, which contains sentence-aligned texts in 21 European languages that have been extracted from the proceeding of the European parliament. 
To prepare the dataset, we parsed the English sentences with the CLaC discourse parser~\cite{laali16} to identify English DCs and the discourse relation that they signal. The CLaC parser has been learned on Section 02-20 of the PDTB and can disambiguate the usage of the 100 English DCs listed in the PDTB with an F1-score of 0.90 and label them with their PDTB discourse relation with an F1-score of 0.76 when tested on the blind test set of the CoNLL 2016 shared task~\cite{xue16}. This parser was used because its performance is very close to that of the state of the art~\cite{oepen16} (i.e. 0.91 and 0.77 respectively), but is more efficient at running time than~\citet{oepen16}.

Note that since the CoNLL~2016 blind test set was extracted from Wikipedia and its domain and genre differ significantly from the PDTB, the 0.90 and 0.76 F1-scores of the CLaC parser can be considered as an estimation of its performance on texts with a different domain/genre such as Europarl.

\subsection{Mapping Discourse Relations}
\label{sec:build-dictionaries}

To label French DCs with a PDTB discourse relation, we assumed that if a French DC is aligned to an English DC tagged with a discourse relation \textit{Rel}, then it should signal the same discourse relation \textit{Rel}. For our experiment, we used the inventory of 100 English DCs from the PDTB~\cite{prasad08} and the 371 French DCs from LEXCONN V2.1~\cite{danlos15}. For the mapping, we used the subset of 14 PDTB discourse relations that was used in the CoNLL shared task~\cite{xue15}. This list is based on the second-level types and a selected number of third-level subtypes of the PDTB discourse relations. 

% To have statistically reliable results, we ignored 11\% of the LEXCONN French DCs that appeared less than 50 times in Europarl. Therefore in our analysis, we used 89\% of the LEXCONN French DCs which have a frequency higher than 50. 

To have statistically reliable results, we ignored French DCs that appeared less than 50 times in Europarl. Out of the 371 French DCs listed in LEXCONN, seven do not appear in Europarl and 55 have a frequency lower than 50. This means that 89\% (309/371) of the French DCs have a frequency higher than 50 and were thus used in the analysis. A manual inspection of the infrequent DCs shows that they are either informal (e.g. \textit{des fois que}) or rare expression (e.g. \textit{en dépit que}).  Table~\ref{tbl:fr-dc-freq} shows the distribution of the LEXCONN French DCs in Europarl.

\begin{table}[ht]
\begin{center}
\begin{tabular}{|r|rrrr|} \hline 
\rowcolor{LightGray}
    \textbf{Freq.}     &   $\mathbf{= 0}$ & $\mathbf{\le 50}$ & $\mathbf{> 50}$ & \textbf{Total} \\ \hline 
    \# FR-DC & 7 & 55 & 309 & 371 \\ \hline
\end{tabular}
\end{center}
\caption[.]{Distribution of LEXCONN French DCs in the Europarl corpus.}
\label{tbl:fr-dc-freq}
\end{table}

%\vspace{-.3cm}
We used the Moses statistical machine translation system~\cite{koehn07} to extract the number of alignments between French DCs and English DCs. As part of its translation model, Moses generates a phrase table (see Table~\ref{tbl:phrase-table}) which aligns phrases between the language pairs. The phrase table is constructed based on statistical word alignment models and contains the frequency of the alignments between phrase pairs. We used the~\citet{och03} heuristic and combined IBM Model~4 word alignments~\cite{brown93} to construct the phrase table.

Because an English DC can signal different discourse relations, to ensure that Moses's phrase table distinguishes the different usages of the same English DC, we modified its English tokenizer so that each English DC and its discourse relation make up a single token. For example, the token `\textit{although}\textsc{-Concession}' will be created for the DC \textit{although} when it signals the discourse relation \textsc{Concession}. Table~\ref{tbl:phrase-table} shows a few entries of the phrase table for the French DC \textit{même si}. As the table shows, \textit{même si} was aligned to three English DCs: \textit{although}, labeled by the CLaC parser  as a \textsc{Contrast} or as a \textsc{Concession} and to \textit{even if} and \textit{even though} which were not tagged .

\begin{table}[ht]
\centering
\resizebox{\columnwidth}{!}{
\rowcolors{2}{LightGray}{white}
\begin{tabularx}{1\columnwidth}{|XXlr|}
\hline
\rowcolor{Gray}
\textbf{FR-DC} & \textbf{EN-DC} & \textbf{Relation} & \textbf{Freq}  
\begin{filecontents*}{phrase-table.csv}
Fr, En, Rel, Freq
même si, even if, - , 2538
même si, even though, - , 1895
même si, although, Contrast, 1446
même si, although, Concession, 858
\end{filecontents*}
\csvreader[head to column names]{phrase-table.csv}{}
{\\ \hline \textit{\Fr} & \textit{\En} & \textsc{\Rel} & \Freq}
\\ \hline
\end{tabularx}}

\caption[.]{A few entries of the phrase table for the connective \textit{même si}. }
\label{tbl:phrase-table}
\end{table}

% \vspace*{-.2cm}
In total, 1,970 entries of the phrase table contained a French DC, an English DC and a discourse relation\footnote{We only considered entries whose texts are an exact match of an English DC listed in the PDTB and a French DC listed in LEXCONN.}. From these, we computed the number of times a French DC was aligned to each discourse relation, then, created \dict: tuples of \textless \textit{FR-DC, Rel, Prob}\textgreater, where \textit{FR-DC} and \textit{Rel} indicate a French DC and a discourse relation and \textit{Prob} indicates the probability that \textit{FR-DC} signals \textit{Rel}. To calculate \textit{Prob}, we divided the number of times \textit{FR-DC} is associated to \textit{Rel} by the frequency of \textit{FR-DC} in Europarl. In total, the approach generated a lexicon of 900 such tuples, a few of which are shown in~Table~\ref{tab:tuples}.

\begin{table}[ht]
\centering
\resizebox{\columnwidth}{!}{
\rowcolors{2}{LightGray}{white}
\begin{tabularx}{1\columnwidth}{|XXr|}
\hline
\rowcolor{Gray}
\textbf{FR-DC} & \textbf{Relation} & \textbf{Prob}
\begin{filecontents*}{tuples.csv}
Rel, DC, P
Condition, si, 0.27
Concession, même si, 0.08
Condition, lorsque, 0.05
Concession, néanmoins, 0.07
\end{filecontents*}
\csvreader[head to column names]{tuples.csv}{}
{\\ \hline \textit{\DC} & \textsc{\Rel} & \P}
\\ \hline
\end{tabularx}}
\caption[.]{A few entries of \dict.
 %of \textless \textit{FR-DC, Rel Prob}\textgreater
 }
\label{tab:tuples}
\end{table}

\section{Evaluation}
\label{sec:evaluation}

To evaluate \dict, because LEXCONN uses a different inventory of discourse relations than the PDTB, we only considered the discourse relations that are common across these inventories: \textsc{Concession} and \textsc{Condition}. According to LEXCONN, 61 French DCs can signal a \textsc{Concession} or a \textsc{Condition} discourse relation. Out of these, 44 have a frequency higher than 50 in Europarl. 

\subsection{Automatic Evaluation}

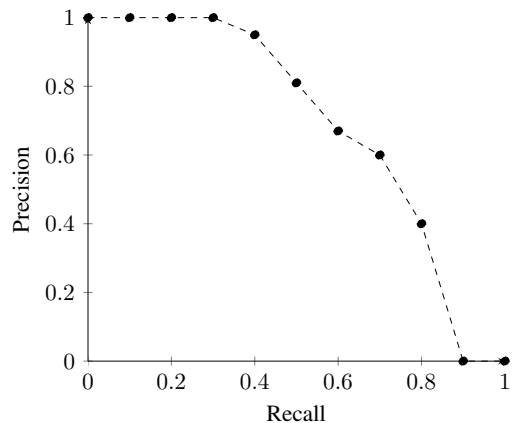
\begin{figure}[!ht]
    \centering
    \begin{tikzpicture}[scale=0.8]
        \begin{axis}[
            xlabel=Recall,
            ylabel=Precision,
            xmin=0,
            ymin=0,
            xmax=1,
            ymax=1,
            axis x line=bottom,
            axis y line=left,
            %xlabel style={bottom},
            % ylable style={left},
            ]
            \addplot[color=black,mark=*,dashed]  table [x=R, y=P, col sep=comma] {by-def-en.csv};
            %\addlegendentry{Raw}
            %\addplot[color=black,mark=*,dotted]  table [x=R, y=P, col sep=comma] {by-def-enfr.csv};
            %\addlegendentry{Parsed}
        \end{axis}
    \end{tikzpicture}
    \caption{11-Point Interpolated Average Precision Curve.}
    \label{fig:curve-map-by-definition}
\end{figure}

To measure the quality of \dict, we ranked the \textless \textit{FR-DC, Rel, Prob}\textgreater~tuples based on their probability and measured the quality of the ranked list using 11-point interpolated average precision~\cite{manning08}. 
This curve shows the highest precision at the 11 recall levels of 0.0, 0.1, 0.2, ..., 1.0. 
This method allows us to evaluate the ranked list without considering any arbitrary cut-off point. As Figure~\ref{fig:curve-map-by-definition} shows, the approach retrieved 50\% of the French DCs in LEXCONN with a precision of 0.81.

In addition, we also computed Average Precision (AveP)~\cite{manning08}; the average of the precision obtained after seeing a correct LEXCONN entry in \dict. 
More specifically, given a list of ranked tuples:
\begin{equation}
    AveP=\frac{1}{N}\sum_{i=1}^{N}Precision(DC_i)
\end{equation}
where $N$ is the number of LEXCONN French DCs that signals the \textsc{Concession} or \textsc{Condition} discourse relations (i.e. 44), $DC_i$ is the rank of the $i^{th}$ LEXCONN DC in \dict, and $Precision(DC_i)$ is the precision at the rank $DC_i$ of the ranked tuples. 
It can be shown that $AveP$ approximates the area under the interpolated precision-recall curve~\cite{manning08}. 
The proposed approach identified 36 (81\%) of these 44 French DCs with an $AveP$ of 0.68.

\begin{table*}[t]
            \centering
            \resizebox{\textwidth}{!}{
            \rowcolors{2}{LightGray}{white}
            \begin{tabularx}{\textwidth}{|X|l|c||X|l|c|}
            \hline
            \rowcolor{Gray}
             \textbf{FR-DC} & \textbf{Relation} & \textbf{Jdg} & \textbf{FR-DC} & \textbf{Relation} & \textbf{Jdg}
            \begin{filecontents*}{manual-judge2.csv}
ID,Rel,DC,Prob,Freq,judge,trans,transInSense,Ida,Rela,Dca,Proba,Freqa,judgea,transa,transInSensea
1,condition,à défaut de,0.03,320,\checkmark,unless,if,8,concession,tout de même,0.02,1665,\checkmark,nevertheless,nonetheless
2,concession,cependant,0.01,21413,\checkmark,however,nonetheless,9,concession,toutefois,0.01,28291,\checkmark,however,nonetheless
3,condition,faute de,0.03,870,\checkmark,otherwise,if,10,condition,pour autant,0.05,1699,$\times$,however,if
4,concession,malgré tout,0.05,1214,\checkmark,nevertheless,nonetheless,11,condition,sinon,0.03,2052,$\times$,otherwise,if
5,concession,néanmoins,0.07,10309,\checkmark,however,nonetheless,12,concession,certes,0.01,5325,$\times$,however,although
6,concession,nonobstant,0.02,185,\checkmark,however,although,13,condition,lorsque,0.05,28507,$\times$,when,if
7,concession,quand même,0.03,1574,\checkmark,nevertheless,nonetheless,14,condition,pour que,0.02,17469,$\times$,so that,if
\end{filecontents*}
\csvreader[head to column names]{manual-judge2.csv}{}
            {\\\hline  \textit{\DC/\transInSense} & \textsc{\Rel} & \judge & \textit{\Dca/\transInSensea} & \textsc{\Rela} & \judgea}
            \\ \hline
            \end{tabularx}
            }
        \caption{Error analysis of the potential false positive entries. \checkmark indicates newly discoursed mappings which are not included in LEXCONN.
        %The last column indicates the human judgment. 
        %True false-positives are marked by $\times$.
        }
    \label{tab:true-positive}
\end{table*}%

\subsection{Manual Evaluation}

In addition to the quantitative evaluation, we also performed a manual analysis of the false-positive errors to see if they really constituted errors.  To do so, we looked at the tuples with a probability higher than 0.01 but which did not appear in LEXCONN. 14 such cases, shown in Table~\ref{tab:true-positive}, were found.  

For example, while the French connective \textit{à défaut de} (\#1 in Table~\ref{tab:true-positive}) signals a \textsc{Condition} discourse relation in Sentence~\ref{ex:condition} below, only the \textsc{Explanation} and the \textsc{Concession} discourse relations were associated with this connective in LEXCONN. 

\begin{enumerate}[label*=(\arabic*)]
    \item \label{ex:condition} 
    \textbf{FR:} \underline{À défaut de} se montrer très ambitieux, notre industrie, nos chercheurs et nos experts ne disposeront purement et simplement pas du brevet moderne dont ils ont besoin. \\
    \textbf{EN:} \underline{If} we are anything less than ambitious in this field, we shall simply not provide our industry, our research and development experts with the modern patent which they need.  
\end{enumerate}

To evaluate if these 14 cases were true mistakes, we randomly selected five English-French parallel sentences from Europarl that contained the French DC and one of its English DC translations signalling the discourse relation. Then, we showed the French DCs within their sentence to two native French speakers and asked them to confirm if the discourse relation identified was indeed signaled  by the French DCs or not. The Kappa agreement between the two annotators was 0.72.
For 9 French connectives, both annotators agreed that 
indicated that in at least one of the five sentences, 
the discourse relation was signalled by the connective.  This indicates that 64\% (9/14) are in fact true-positives, i.e. correct mappings that are not listed in LEXCONN.  Table~\ref{tab:true-positive} shows the 14 pairs of \textless FR-DC/English translation, Discourse relation\textgreater\ used in the manual evaluation and indicates the newly discovered mappings by \checkmark.

% \begin{table}[ht]
%     \centering
%     \resizebox{\columnwidth}{!}{
%     \rowcolors{2}{LightGray}{white}
%     \begin{tabularx}{1.22\columnwidth}{|X|l|c|}
%     \hline
%     \rowcolor{Gray}
%      \textbf{FR-DC/En Translation} & \textbf{DR} & \textbf{Jdg}%& \textbf{\boldmath $P(R^L|C^F)$} & \boldmath $f(C^F)$
%     \csvreader[head to column names]{manual-judge.csv}{}
%     {\\\hline \textit{\DC}/\textit{\trans} & \textsc{\Rel} & \judge }% & \Prob & \Freq }
%     \\ \hline
%     \end{tabularx}}
%     \caption{Error analysis of the potential false positive entries. 
%     %The last column indicates the human judgment. 
%     %True false-positives are marked by $\times$.
%     }
%     \label{tab:true-positive}
% \end{table}

We also observed that if multiple explicit connectives occur in the same clause (e.g. \textit{certes} and \textit{mais}), one of them can affect the discourse relation signaled by the other. This is an interesting phenomenon as it seems to indicate that the connectives are not independent. For example, in Sentence~\ref{ex:dc-dependent}, the combination of \textit{certes} and \textit{mais} signals a \textsc{Concession} discourse relation.

\begin{enumerate}[label*=(\arabic*),resume]
    \item \label{ex:dc-dependent} 
    \textbf{FR:} Cela coûte \underline{certes} un peu plus cher, \underline{mais} est sans conséquence pour l'environnement. \\
    \textbf{EN:} \underline{Although} it is a little more expensive, it does not harm the environment.
\end{enumerate}

Note that according to LEXCONN, neither \textit{certes} nor \textit{mais} can signal a \textsc{Concession} discourse relation. The same phenomenon was also reported in the PDTB corpus \cite[p. 5]{prasad08-a}.

\section{Conclusion and Future Work}
\label{sec:conclusion}

In this paper, we proposed a novel approach to automatically map PDTB discourse relations to French DCs. Using this approach, we generated \dict: a lexicon of French DCs and their PDTB discourse relations. When compared with LEXCONN, our approach achieved a recall of 0.81 and an Average Precision of 0.68 for the \textsc{Concession} and \textsc{Condition} discourse relations. A manual error analysis of the false-positives showed that the approach identified new discourse relations for 9 French DCs which are not included in LEXCONN. As future work, we plan to evaluate all the discourse relations in \dict\ and apply the approach to other languages.

\subsubsection* {Acknowledgement}
The authors would like to thank the anonymous referees for their insightful
comments on an earlier version of the paper. Many thanks also to Andre Cianflone for his help on the evaluation of this work. This work was financially supported by an NSERC grant.

\end{document}